\documentclass[letterpaper, 10pt, conference]{ieeeconf}

\IEEEoverridecommandlockouts
\usepackage{cite}
\usepackage{amsmath,amsfonts}
\usepackage{graphicx}
\usepackage{booktabs}
\usepackage{bm}
\usepackage{flushend}
\title{\LARGE 
Vision-Based Control of a Tether-Suspended Aerial Radiation Sensing Payload
}

\author{Ian Snider$^{1}$, Brian J. Quiter$^{2}$, Emil Rofors$^{2}$, and Mark W. Mueller$^{3}$%
\thanks{$^{1}$Department of Nuclear Engineering, University of California, Berkeley, CA 94720, USA}%
\thanks{$^{2}$Lawrence Berkeley National Laboratory, Berkeley, CA 94720, USA}%
\thanks{$^{3}$Department of Mechanical Engineering, University of California, Berkeley, CA 94720, USA}%
}

\begin{document}

\maketitle

\begin{abstract}
Aerial radiation surveys achieve higher sensitivity when the radiation detector is held close to the ground. Detector sensitivity falls off roughly with the inverse square of the distance to the source, so a detector flown high is slower to reach a given minimum detectable activity. Flying the vehicle low puts the propellers near the ground, where downwash can disturb the surveyed area and resuspend contaminated particulates. Tether suspension decouples the detector from the vehicle altitude, but leaves the payload unactuated and only indirectly controllable. We therefore present a vision-based control approach for an aerial sensing payload suspended on a tether beneath a heavy-lift drone. Because a survey plan is decided as radiation detections arrive, we design a pilot aid for commanding the survey trajectory manually with a handheld transmitter. The controller regulates the payload, rather than the vehicle, onto that trajectory. The system uses onboard sensors with a downward-facing camera fixed to the drone body tracking a ring marker on the payload. A four-state Kalman filter estimates the tether swing angles and rates from payload bearing measurements, and a linear quadratic regulator with integral action takes the payload position as the regulated output. In outdoor flight tests under wind, the payload-aware controller reduced payload tracking error during transit by 20\% when compared against a vehicle-referenced baseline, with the cost of higher peak error on arrival at a waypoint.
\end{abstract}

\section{Introduction}

Robotic surveys of harmful radiation reduce the risk posed to humans when characterizing contamination at disaster and clean-up sites. Unmanned Aerial Vehicles (UAVs) have increasingly been used to survey radioactive contamination and enable rapid characterization of a large site. Following the Fukushima Dai-ichi accident, both unmanned helicopters and multirotors were deployed to map radiation dose rates and identify hotspots over contaminated terrain \cite{radiation_towler_2012, aerial_sanada_2014, first_mochizuki_2017, remote_sato_2018}, with payloads including gamma-imaging systems \cite{prototype_jiang_2016} and Compton cameras that identify source direction and intensity \cite{remote_sato_2018}.\par

The primary limitation on survey quality is source proximity. Detector sensitivity is roughly proportional to the inverse square of the distance to the radioactive source; therefore, high-resolution surveys are flown close to the ground. At these altitudes, vehicles must adapt to complex terrain contours and risk contamination, and rotor downwash can resuspend contaminated particulates in the soil \cite{limiting_ravehamit_2022}. Tether-suspension keeps the vehicle at a safe altitude relative to cluttered or contaminated terrain. However, attaching a pendulum to an aircraft creates complex flight behavior that most UAVs are not readily able to accommodate.\par

\begin{figure}[t]
    \centering
    \vskip 5pt
    \includegraphics[width=\columnwidth]{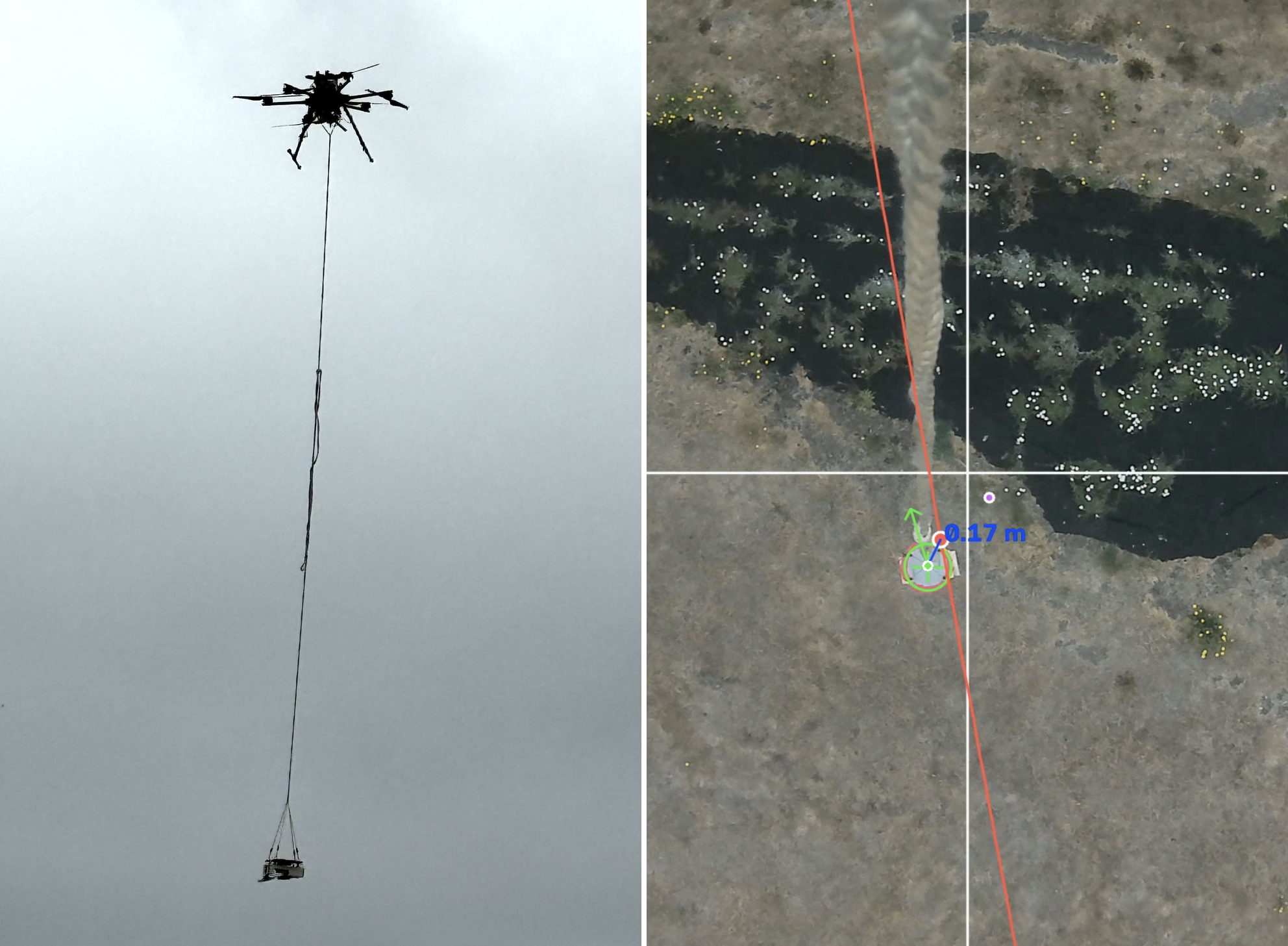}
    \caption{Heavy-lift drone flying with tether-suspended payload. The right image is the downward-facing camera view with the overlaid reference trajectory: the current reference point (red), the payload position estimate (green), a line giving the estimated distance from the reference (blue), and the point at which a vertically hung tether would sit beneath the drone at hover (purple).}
    \label{fig:robot}
\end{figure}

 The tether-suspension problem has been studied extensively in aerial transport literature \cite{review_estvez_2024,survey_villa_2020, suspended_klausen_2014, foehn_fast_2017}, where the objective is usually cargo delivery and the suspended payload is a disturbance to reject. Transport approaches include swing-free trajectory generation \cite{swing_free_palunko}, deliberately exploiting large swings during periods with zero cable tension \cite{mixed_tang_2015}, and geometric control of the hybrid system \cite{dynamics_sreenath_2013}. In this work, the control objective is to regulate the payload's position rather than the drone's. When the trajectory objective, the payload state must be estimated. Laboratory systems typically use motion capture \cite{survey_villa_2020}, while field systems must rely on the drone's onboard sensors.  A vision-based approach provides measurements of the payload state using a downward-facing camera and avoids the infrastructure needed to power the payload \cite{vision_slabber_2021, dynamic_gassner_2017, aggressive_tang_2018}. Our estimation architecture follows the vision-based approach of Tang et al. \cite{aggressive_tang_2018}, demonstrated for agile control of payloads on tethers up to 0.7~m. We adopt the estimator but not the aggressive flight regime: our heavy-lift platform and substantially longer tether make such maneuvers undesirable when personnel are nearby. In the less agile regime, the dominant disturbance is wind acting on the tether and payload. Existing treatments are primarily computational. Path-following control \cite{pathfollowing_qian_2020} and sliding-mode control of a heavy-lift quadrotor with suspended payload \cite{modeling_gomiero_2024} are evaluated in simulation, and experimental validation of tethered payload control is predominantly indoors. The effect of wind on the outdoor heavy-lift tethered payload system remains largely uncharacterized. Our main contributions are an outer-loop controller that regulates payload position through acceleration commands to an unmodified ArduPilot inner loop, a payload bearing estimator that recovers the ring center from partial arcs under continuous tether occlusion, a reference generator that allows a pilot to command the payload rather than the vehicle, and outdoor flight tests that evaluate the payload-aware approach against a vehicle-referenced baseline.

\section{Modeling}\label{sec:modeling}
\begin{figure}[t]
    \centering
    \vskip 5pt
    \includegraphics[width=\columnwidth]{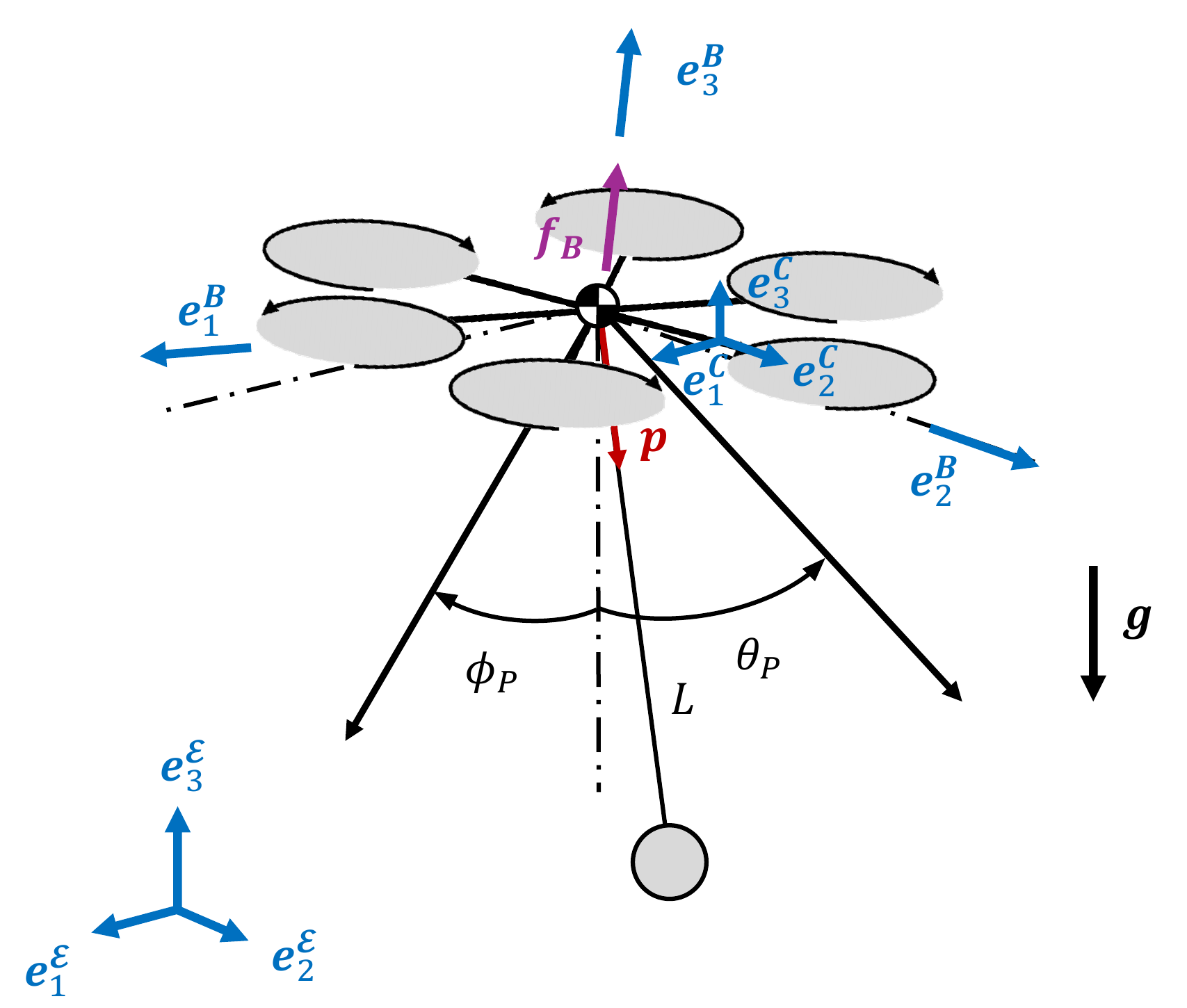}
    \caption{Diagram of drone with tether-suspended payload.}
    \label{fig:model}
\end{figure}

We consider the drone vehicle and payload as two point masses, $m_B$ and $m_P$, joined by a massless tether of length $L$. Fig. \ref{fig:model} contains a schematic of the drone with tether-suspended payload in the inertial earth-fixed frame $\mathcal{E}$.
The body frame of the drone is denoted with $\mathcal{B}$, and the camera frame is denoted with $\mathcal{C}$. A superscript denotes the frame in which a vector is coordinated. An arbitrary frame $\mathcal{A}$ has elementary unit vectors $\{\bm{e}_1^\mathcal{A}, \bm{e}_2^\mathcal{A}, \bm{e}_3^\mathcal{A}\}$, with $\bm{e}_3^\mathcal{E}$ directed upward.\par
We use the usual approach of defining a unit vector $\bm{p}$ from the drone to the payload \cite{trajectory_sreenath_2013}. Because the drone is heavy and must operate with personnel nearby, aggressive swings are neither needed nor desired, and a manual override is preferred over an autonomous maneuver at large swing angles. We therefore restrict our analysis to small angles, allowing linear approximation of the nonlinear unit vector:
\begin{equation}\label{eq:unit_vector}
    \bm{p}^\mathcal{E} \approx \begin{bmatrix}
       \phi_P & \theta_P & -1
    \end{bmatrix}^\top
\end{equation}
where $\phi_P$ and $\theta_P$ are the swing angles of the pendulum. Assuming the tether remains taut, the payload and vehicle positions, $\bm{s}_P$ and $\bm{s}_B$, satisfy the rigid constraint $\bm{s}_P = \bm{s}_B + L\bm{p}$, and the tether carries tension $T \geq 0$ acting on the payload along $-\bm{p}$ and on the vehicle along $+\bm{p}$, giving the standard model of a drone with a tether-suspended load \cite{trajectory_sreenath_2013, geometric_goodarzi_2015, modeling_gomiero_2024}
\begin{align}
    m_P \ddot{\bm{s}}_P &= -T \bm{p} - m_P \bm{g}
        \label{eq:payload_dynamics} \\
    m_B \ddot{\bm{s}}_B &= \bm{f}_B + T\bm{p} - m_B \bm{g}
        \label{eq:drone_dynamics}
\end{align}

where $\bm{f}_B$ is the thrust from the propellers, $g$ is the gravitational acceleration, and $\bm{g} = g\bm{e}_3^\mathcal{E}$. We define $\bm{a}_B = \ddot{\bm{s}}_B + \bm{g}$, the accelerometer specific force, where $\bm{a}_B^\mathcal{E} = \begin{bmatrix}a_1 & a_2 & a_3\end{bmatrix}^\top$. For the controller and payload estimator of Sec.~\ref{sec:control} and Sec.~\ref{sec:estimation} we need only the pendulum dynamics and may decouple the drone dynamics. Differentiating the constraint twice, eliminating $T$ from the vertical row of \eqref{eq:payload_dynamics}, and linearizing about hover, where $a_3 \approx g$, gives two decoupled pendulum equations:
\begin{align}
    \ddot{\phi}_P &= -\frac{g}{L}\phi_P - \frac{1}{L}a_1 \label{eq:alpha1} \\
    \ddot{\theta}_P &= -\frac{g}{L}\theta_P - \frac{1}{L}a_2 \label{eq:alpha2}
\end{align} 
The vehicle attitude and thrust dynamics of \eqref{eq:drone_dynamics} are not required for the outer loop controller design of Sec. \ref{sec:control}; the outer loop models the vehicle as a double integrator driven by  $\bm{a}_B$. The tension reaction on the vehicle is rejected as a disturbance by the inner attitude and thrust loops of the ArduPilot flight controller. Damping is not included in the model, so the estimator of Sec. \ref{sec:estimation} absorbs the aerodynamic drag into its process noise.

\section{Controller Design}\label{sec:control}

\begin{figure}[t]
    \centering
    \vskip 5pt
    \includegraphics[width=\columnwidth]{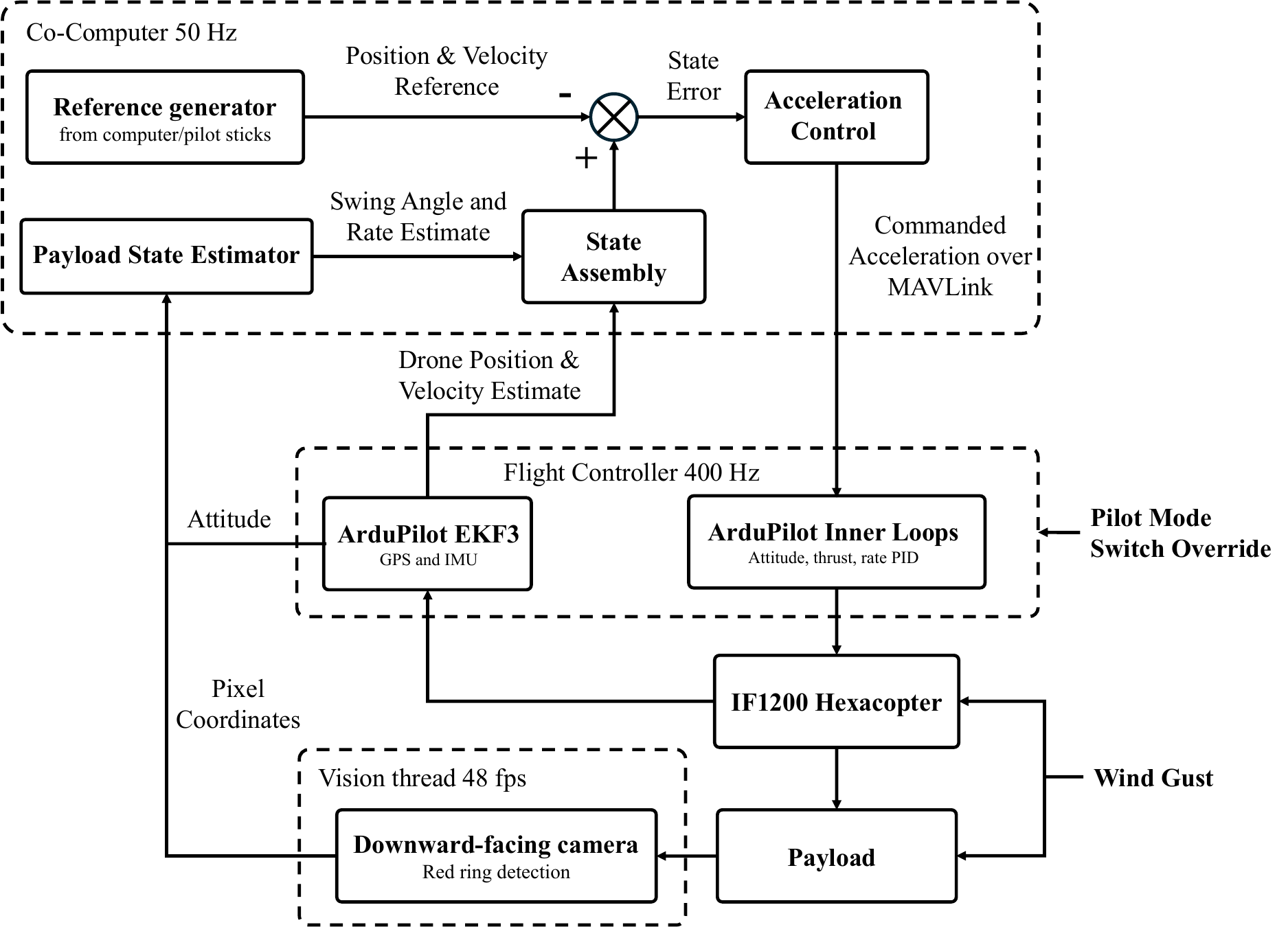}
    \caption{Block diagram of the control architecture. The system is separated into the 50 Hz payload estimator and acceleration commands, the ArduPilot inner-loop flight controller, and the 48 fps vision thread. The system receives external input from the reference generator, pilot, and wind.}
    \label{fig:block_diagram}
\end{figure}

The goal of the designed controller is to regulate the position of the payload to a desired setpoint. For safety, we limit the drone's behavior by building the controller around the existing ArduPilot infrastructure. The control system  then sends acceleration setpoints over MAVLink to the ArduPilot inner-loop controller. A block diagram of the control architecture is shown in Fig. \ref{fig:block_diagram}.

\subsection{ArduPilot  Inner-Loop} \label{sec:ardupilot}
The ArduPilot inner-loop attitude control is preserved as a failsafe for the experimental outer-loop controller. The ArduPilot inner-loop system is configured so that it is only able to receive external outer-loop commands when in \verb|GUIDED| mode. The pilot may recover the vehicle at any time by switching out of \verb|GUIDED| via a transmitter or ground control station (GCS). Acceleration setpoints are streamed at 50 Hz as \verb|SET_POSITION_TARGET_LOCAL_NED| messages with every field masked except the acceleration and yaw. The ArduPilot position and velocity error correction is disabled so that the acceleration command enters the inner loop as a feedforward term. The only shaping ArduPilot applies to the command is a 10 m/s$^3$ jerk limit. In the outer-loop design of Sec. \ref{sec:acceleration}, the inner-loop is treated as an ideal accelerator since its bandwidth is above the pendulum swing frequency.

\subsection{Acceleration Controller} \label{sec:acceleration}

We design a linear-quadratic-regulator (LQR) with integral action (LQI) to regulate the payload error and reject wind. We start by building an LQR controller with the system states in $\bm{x} \in \mathbb{R}^{10}$ and command input $\bm{u} \in \mathbb{R}^3$

\begin{equation}
    \label{eq:states}
    \bm{x} = \begin{bmatrix}
        {\bm{s}_B^\mathcal{E}}^\top & {\bm{v}_B^\mathcal{E}}^\top & {\bm{\xi}^\mathcal{E}}^\top
    \end{bmatrix}^\top, \qquad \bm{u} = \bm{a}_B^\mathcal{E}
\end{equation}
where $\bm{s}_B^\mathcal{E}$ and $\bm{v}_B^\mathcal{E}$ are the vehicle position and velocity, and $\bm{\xi}^\mathcal{E}$ is the payload state in $\mathcal{E}$
\begin{equation}
    \label{eq:kf_state}
    \bm{\xi}^\mathcal{E} = \begin{bmatrix}
    \phi_P & \theta_P & \dot{\phi}_P & \dot{\theta}_P
    \end{bmatrix}^\top \in \mathbb{R}^4
\end{equation}
The $A$ and $B$ matrices of the state-space model, $\dot{\bm{x}} = A \bm{x} + B \bm{u}$, follow by assembling the vehicle double integrator, $\dot{\bm{s}}_B^\mathcal{E} = \bm{v}_B^\mathcal{E}$ and $\dot{\bm{v}}_B^\mathcal{E} = \bm{a}_B^\mathcal{E}$, with the linearized pendulum dynamics of (\ref{eq:alpha1}) and (\ref{eq:alpha2}). The quantity we wish to control is the payload position, which is output from the state as $\bm{y} = C \bm{x}$ with

\begin{equation}
    \label{eq:output}
    \bm{y} = \begin{bmatrix}
        s_{B,1}^\mathcal{E} + L \phi_P \\
        s_{B,2}^\mathcal{E} + L \theta_P \\
        s_{B,3}^\mathcal{E}
    \end{bmatrix}
\end{equation}
and $C \in \mathbb{R}^{3 \times 10}$ follows by inspection. The constant vertical offset of length $L$ is a component of the equilibrium condition rather than the output. The equilibrium is expressed as
\begin{equation}
    \label{eq:equilibrium}
    \bm{x}^* = \begin{bmatrix}
    \bm{s}_{P,\text{ref}}^\mathcal{E} + L \bm{e}_3^\mathcal{E} \\
    \dot{\bm{s}}_{P, \text{ref}}^\mathcal{E} \\
    \bm{0}_{4 \times 1}
    \end{bmatrix}
\end{equation}

Under persistent disturbances, such as wind, proportional state feedback leaves a steady-state payload offset. We therefore augment the system with the integral output error, $\bm{e}_I = \int (\bm{y} - \bm{y}_\text{ref}) \text{d}t$, where $\bm{y}_\text{ref} = C \bm{x}^*$. The state-space of the augmented system with state $\bm{z}$ is
\begin{align}
    \label{eq:augmented}
   \dot{\bm{z}} = \overline{A}\bm{z} + \overline{B}\bm{u}, \qquad
    \bm{z} = \begin{bmatrix} \bm{e}^\top & \bm{e}^\top_I \end{bmatrix}^\top
    \in \mathbb{R}^{13}
\end{align}
where $\bm{e} = \bm{x} - \bm{x}^*$ and the state and input matrices become,
\begin{equation}
    \overline{A} = \begin{bmatrix} A & \bm{0} \\ C & \bm{0} \end{bmatrix}, \qquad
    \overline{B} = \begin{bmatrix} B \\ \bm{0} \end{bmatrix}
\end{equation} 

The feedback controller gain matrix $\overline{K}$ depends only on the tether length $L$ and LQI cost with the augmented state weighting matrix $\overline{Q}$ and input weighting matrix $R$. The infinite-horizon cost $J$ is
\begin{equation}
    \label{eq:lqi_horizon}
    J = \int_0^\infty \left(\bm{z}^\top \overline{Q} \bm{z} + \bm{u}^\top R \bm{u} \right) \text{d}t
\end{equation}

where $\overline{Q}$ and $R$ are constructed as

\begin{equation}
    \overline{Q} = \begin{bmatrix}
        C^\top W C & 0_{10 \times 3} \\
        0_{3 \times 10} & W_I
    \end{bmatrix}, \qquad R = \rho I_{3 \times 3}
\end{equation}

and $W = \text{diag}(w_x, w_y, w_z)$ weights the payload position error, $W_I = \text{diag}(w_{x,I}, w_{y,I}, w_{z,I})$ weights the integral, and $\rho$ is a control effort tuning constant. We set $w_x = w_y$ and $w_{x,I} = w_{y,I}$ by the horizontal symmetry of the dynamics. Solving the algebraic Riccati equation for $\overline{Q}$ and $R$ yields the gain $\overline{K} = [\, K_x \;\; K_I \,]$, where $K_x = [\, K_s \;\; K_v \;\; K_{\phi\theta} \;\; K_{\dot\phi\dot\theta} \,]$ with gains on vehicle position, vehicle velocity, swing angle, and swing rate. The control input is
\begin{equation}
\bm{u} = -\begin{bmatrix}
    K_x & K_I
\end{bmatrix} \begin{bmatrix}
    \bm{e} \\ \bm{e}_I
\end{bmatrix}
\end{equation}
 The above formulation of the LQI controller cannot guarantee stability when the drone actuators saturate. Integral action thus accumulates error whenever the commanded acceleration cannot be achieved. To prevent integral wind-up we implement a conditional integration
\begin{equation}
    \label{eq:antiwindup}
    \dot{\bm{e}}_I = \begin{cases}
        \bm{y} - \bm{y}_\text{ref}, & \lVert \bm{y} - \bm{y}_\text{ref} \rVert < \mu
        \;\text{ and }\; \lVert K_I \bm{e}_I \rVert \leq u_{I,\max} \\[1ex]
        \bm{0}, & \text{otherwise}
    \end{cases}
\end{equation}
Integration is enabled only inside an error band $\mu$ and the second condition caps the integral command at $u_{I,\max}$. 

\subsection{Pilot Aid} \label{sec:pilot}
 The reference for the controller is a position and velocity setpoint moving along the reference trajectory. Users of the system often prefer to have manual control over the payload. To meet this objective, we read signals from the sticks of a transmitter. The stick signals are normalized and multiplied by a preset maximum payload velocity. The commanded velocity is bounded with an acceleration limit to ramp the reference velocity, and integration then yields the position setpoint.\par

In practice, integral action contributes to large swings that should be avoided during manual control, so an LQR may be preferred when the pilot cannot see the generated reference. Under approximately constant wind the payload holds a constant offset that the pilot compensates through the sticks. However, we still use LQI so that we can analyze the reference tracking capabilities.

\section{Payload Estimation} \label{sec:estimation}
Accurate payload state information is required for the controller of Sec. \ref{sec:control}. A downward-facing camera fixed to the drone body detects a red ring marker on the periphery of the payload body. The camera provides the payload bearing but ignores range. We use a Kalman filter to combine bearing measurements with the measured vehicle acceleration, which enters as a known input. We let $k$ represent the $k$-th discrete time step index. $f(\cdot)$ is the process model with process covariance $\bm{Q}_k$ and $h(\cdot)$ is the measurement model with measurement covariance $\bm{R}_k$. Throughout, $\bm{T}^{\mathcal{AB}}$ denotes the transformation to frame $\mathcal{A}$ from frame $\mathcal{B}$, so that a vector in $\mathcal{B}$ is expressed in $\mathcal{A}$ as $\bm{v}^\mathcal{A} = \bm{T}^{\mathcal{AB}}\bm{v}^\mathcal{B}$.

\subsection{Process Model} \label{sec:process}
A four-state process model is defined with payload swing angles and swing rates. The control input is the measured horizontal acceleration from the vehicle's accelerometer transformed into $\mathcal{E}$
\begin{equation}
    \label{eq:kf_input}
    \bm{u}_k = \begin{bmatrix} a_1 & a_2 \end{bmatrix}^\top, \qquad
    \bm{a}_B^\mathcal{E} = \bm{T}^{\mathcal{EB}}\bm{a}_B^\mathcal{B}
\end{equation}
where $\bm{T}^{\mathcal{EB}}$ is obtained from the ArduPilot attitude estimator. Treating the inertial measurement as an input rather than a measurement keeps the filter state restricted to the payload. The process model is
\begin{equation}
    \label{eq:process_formulation}
    \bm{\xi}_k = f(\bm{\xi}_{k-1}, \bm{u}_{k-1}) + \bm{\varepsilon}_k \qquad
    \bm{\varepsilon}_k \sim \mathcal{N}(\bm{0}, \bm{Q}_k)
\end{equation}
Collecting \eqref{eq:alpha1} and \eqref{eq:alpha2} in state-space form gives the continuous-time dynamics $\dot{\bm{\xi}} = F\bm{\xi} + G\bm{u}$, with
\begin{equation}
    \label{eq:process_matrices}
    F = \begin{bmatrix}
        0 & 0 & 1 & 0 \\
        0 & 0 & 0 & 1 \\
        -\frac{g}{L} & 0 & 0 & 0 \\
        0 & -\frac{g}{L} & 0 & 0
    \end{bmatrix}, \qquad
    G = \begin{bmatrix}
        0 & 0 \\ 0 & 0 \\ -\frac{1}{L} & 0 \\ 0 & -\frac{1}{L}
    \end{bmatrix}
\end{equation}
The state is propagated by Euler integration of \eqref{eq:process_matrices}. Aerodynamic drag on the payload, tether flexing, and wind gusts enter as unmeasured accelerations on the swing rates. We model these as noise entering only the rate states, with intensity $q_{\dot{\phi}}$ and $q_{\dot{\theta}}$, discretized over the prediction interval $\Delta t$,
\begin{equation}
    \label{eq:process_noise}
    \bm{Q}_k = \Delta t\,\text{diag}\!\left(0,\; 0,\; q_{\dot{\phi}},\; q_{\dot{\theta}}\right)
\end{equation}
where the value $q_{\dot{\phi}} = q_{\dot{\theta}} = 0.01 $ was found to work well. The swing states receive no process noise; their uncertainty grows only through the coupling to the rates. 
\subsection{Measurement}
The Kalman filter estimates the swings states of the payload with the payload direction as the measured quantity
\begin{equation}
    \label{eq:measurement_def}
    \bm{\eta}_k = h(\bm{\xi}_k) + \bm{\nu}_k, \qquad
    \bm{\nu}_k \sim \mathcal{N}(\bm{0}, \bm{R}_k)
\end{equation}
where the payload direction predicted from the state is
\begin{equation}
    \label{eq:payload_direction}
    \bm{p}^\mathcal{E}(\bm{\xi}) = \begin{bmatrix} \phi_P & \theta_P & -1 \end{bmatrix}^\top
\end{equation}
The small-angle restriction of Sec.~\ref{sec:modeling} lets the payload
direction be written as \eqref{eq:payload_direction} in $\mathcal{E}$. The
camera, however, reports the bearing in $\mathcal{C}$, so the predicted
direction is rotated into the camera frame before comparison:
\begin{equation}
    \label{eq:camera_rotation}
    \bm{T}_k^{\mathcal{CE}} = \bm{T}^{\mathcal{CB}}\bm{T}^{\mathcal{BE}}, \qquad
    \hat{\bm{p}}^\mathcal{C} = \bm{T}_k^{\mathcal{CE}}\,\bm{p}^\mathcal{E}(\bm{\xi})
\end{equation}
where $\bm{T}^{\mathcal{CB}}$ is the fixed camera transformation, with $\bm{e}_3^\mathcal{C}$ aligned with $-\bm{e}_3^\mathcal{B}$ so that a payload hanging beneath the vehicle has positive depth in $\mathcal{C}$. The camera reports the bearing on the normalized image plane, so the measurement model returns the horizontal components of $\hat{\bm{p}}^\mathcal{C}$ divided by its depth,
\begin{equation}
    \label{eq:measurement_model}
    h(\bm{\xi}_k) = \begin{bmatrix} \hat{p}_1^\mathcal{C} / \hat{p}_3^\mathcal{C} & \hat{p}_2^\mathcal{C} / \hat{p}_3^\mathcal{C} \end{bmatrix}^\top
\end{equation}
Under the small-angle restriction of Sec.~\ref{sec:modeling}, $\hat{p}_3^\mathcal{C} \approx 1$ and the perspective division is neglected, leaving the measurement model affine in $\bm{\xi}_k$. Writing the $ij$th element of $\bm{T}_k^{\mathcal{CE}}$ as $T_{ij}$,
\begin{equation}
    \label{eq:measurement_matrix}
    \bm{H}_k = \begin{bmatrix}
        T_{11} & T_{12} & 0 & 0 \\
        T_{21} & T_{22} & 0 & 0
    \end{bmatrix}, \quad
    \bm{b}_k = -\begin{bmatrix} T_{13} \\ T_{23} \end{bmatrix}
\end{equation}
where $h(\bm{\xi}_k) = \bm{H}_k\bm{\xi}_k + \bm{b}_k$ and offset $\bm{b}_k$ is the camera-view projection of the local vertical and vanishes at hover. The model is affine, so the standard Kalman update applies with innovation $\bm{\eta}_k - (\bm{H}_k\hat{\bm{\xi}}_k^- + \bm{b}_k)$ and $\bm{H}_k$ in the covariance update. The measurement covariance is
\begin{equation}
    \label{eq:measurement_noise}
    \bm{R}_k = \text{diag}(\sigma_x^2, \sigma_y^2)
\end{equation}
where the measurement noise $\sigma_x = \sigma_y = 0.01$ was found to work well. The correction step follows the usual procedure of \cite{probabilistic_thrun_2005}. The prediction step runs at every controller step. The update is applied only on frames for which a measurement is available.

\subsection{Image Processing}\label{sec:image}
\begin{figure}[t]
    \centering
    \vskip 5pt
    \includegraphics[width=0.9\columnwidth]{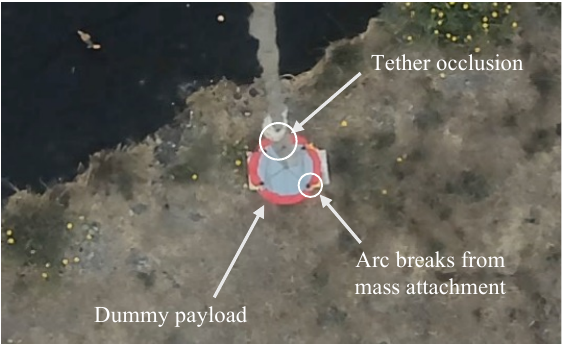}
    \caption{Tether occlusions, as shown in downward-facing camera.}
    \label{fig:tether_occlusions}
\end{figure}

Using OpenCV, we locate the center of a red ring attached to the payload periphery. As shown in Fig. \ref{fig:tether_occlusions}, the tether always partially occludes the red ring.
The ring may also be occluded by equipment attached to the payload. We recover the center of the payload from ring segments by finding the largest visible arc and fit a circle to its pixels. The frame is thresholded in hue-saturation-value (HSV) space to retain pixels within a specified band for red. A morphological closing transformation is applied to close small gaps in the detected arcs. We then apply connected component labeling and evaluate each component. The contours of the component are collected into edge points. Given the $N$ contour points $\{(x_i, y_i)\}_{i=1}^N$, we fit the center $(u_c, v_c)$ and radius $r$ of the detected arc using the algebraic least-squares formulation of Kasa~\cite{circle_kasa_1976}, which linearizes the implicit circle equation by the substitution $c = r^2 - u_c^2 - v_c^2$ and solves the overdetermined system. A $70^\circ$ gate on arc coverage and a fit radius from the tether length helps reject false positives. The fit center is shifted by the fixed camera pivot offset and expressed in pixels. Pixels are converted to a bearing using the calibrated intrinsic matrix $\bm{K}_c$, giving the tether direction in the camera frame
\begin{equation}
    \label{eq:pixel_to_bearing}
    \bm{p}^\mathcal{C} = \bm{K_c}^{-1}\begin{bmatrix} u_c & v_c & 1 \end{bmatrix}^\top
\end{equation}
The measurement taken into the Kalman filter is the first two components of \eqref{eq:pixel_to_bearing}
\begin{equation}
    \label{eq:bearing_measurement}
    \bm{\eta}_k = \begin{bmatrix} p_1^\mathcal{C} & p_2^\mathcal{C} \end{bmatrix}^\top
\end{equation}
Range is not required since the payload is assumed to lie at a fixed distance $L$ along the estimated bearing.

\section{Experiment}
Experiments were conducted on an InspiredFlight IF1200 heavy-lift hexacopter. An ASUS NUC was used for the external control system and payload state estimation. The attached See3CAM\_CU30 camera operated at 2304 $\times$ 1536 pixels with 48 fps. A dummy payload was attached to simulate the radiation detector as shown in Fig. \ref{fig:tether_occlusions}. The physical experiment parameters are collected in Table \ref{tab:physical}.
\begin{table}
\vskip 5pt
\caption{Hexacopter with Load Parameters}
\label{tab:physical}
\centering
\small
\begin{tabular}{ll}
\toprule
Hexacopter mass  & 16.5 kg \\ 
Payload mass  & 5.65 kg \\ 
Tether length  & 6.2 m \\ 
Ring outer diameter  & 31 cm \\ 
Ring width  & 3.5 cm \\ 
\bottomrule
\end{tabular}
\end{table}
\par

For our experiment, we evaluated the payload-aware controller against the ArduPilot vehicle-referenced baseline in outdoor flight under wind. The payload was commanded to fly a square trajectory at 1.25~m/s with a 15~m edge length and 15~s hover at each corner, with an initial 15~s hover at the start. A pre-computed reference trajectory with 2~m/s$^2$ acceleration limit was used instead of the piloting aid for consistency between tests. The test was conducted four times. Two of the tests used the LQI acceleration controller presented in this paper. The LQI costs were tuned using Bryson's rule \cite{bryson1975applied} with tolerances $\tau$ such that the design weight $w = 1/\tau^2$. The controller was set with position tolerances of 1.2~m horizontal and 2.0~m vertical, integral tolerances of 4.0~m$\cdot$s horizontal and 5.0 m$\cdot$s vertical, and $\rho = 3.0$. The control gains were $K_s = \mathrm{diag}(0.85,\,0.85,\,0.58)$, $K_v = \mathrm{diag}(1.68,\,1.68,\,1.07)$, $K_{\phi\theta} = \mathrm{diag}(-3.05,\,-3.05)$, $K_{\dot\phi\dot\theta} = \mathrm{diag}(0.28,\,0.28)$, and $K_I = \mathrm{diag}(0.14,\,0.14,\,0.12)$. Conditional integration used $\mu = 2$~m and $u_{I,\max} = 2$~m/s$^2$. The other two tests used the ArduPilot baseline control system by configuring \verb|MAV_CMD_DO_CHANGE_SPEED| to 1.25~m/s, \verb|WPNAV_ACCEL| to 2~m/s$^2$, and \verb|PSC_JERK_XY|/\verb|PSC_NE_JERK| to 10~m/s$^3$. The drone was manually piloted for takeoff and landing to prevent the tether from tangling. The tests were alternated for fairness under gusty conditions. Wind measurements were collected with a handheld anemometer for each test. Across all four tests, wind came from the south and averaged $\sim$3~m/s with gusts to $\sim$6~m/s.\par

\begin{figure}[t]
    \centering
    \vskip 5pt
    \includegraphics[width=0.9\columnwidth]{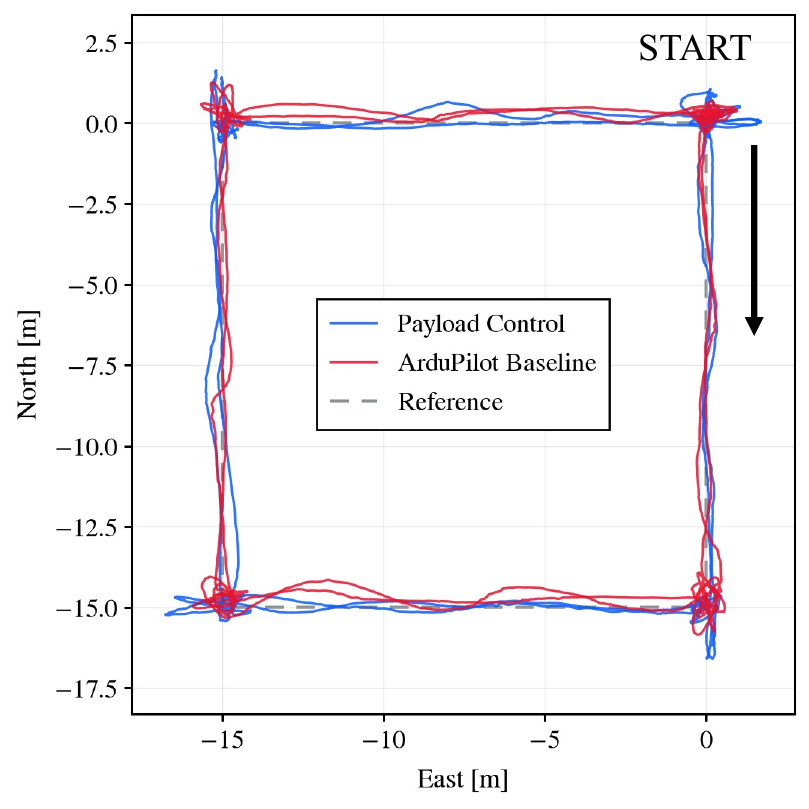}
    \caption{East-North payload trajectories for the two control modes. The test included 15 m legs with 15 s hovers at each corner.}
    \label{fig:payload_trajectory}
\end{figure}

\begin{figure}[t]
    \centering
    \vskip 5pt
    \includegraphics[width=\columnwidth]{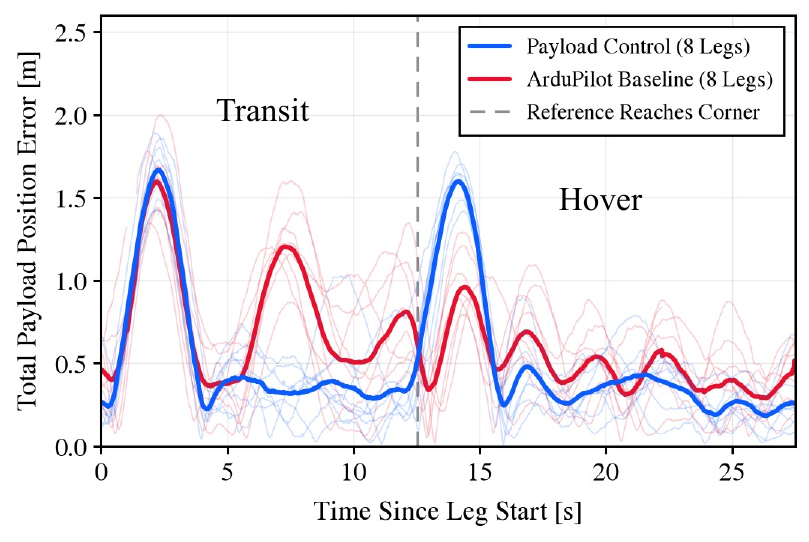}
    \caption{Payload position error versus time, aligned at the start of each leg. The bold blue and red lines are the average payload position error for the payload-aware controller and the vehicle-referenced ArduPilot baseline respectively.}
    \label{fig:position}
\end{figure}

The camera measurement returned a valid ring fit on 98.89\% of 21,172 frames at 48 fps, despite continuous partial occlusion by the tether. The camera-view of the estimated payload bearing is shown in the supplementary video. The longest dropout was 0.44 s, or 9\% of a $2 \pi \sqrt{L/g} = 5.0$~s pendulum period, over which the filter relies on the process model of Sec. \ref{sec:process}. Fig. \ref{fig:payload_trajectory} contains the payload trajectories of the four tests.
Each test lasted 125.5 s. Table \ref{tab:track_error} holds the payload tracking root-mean-square-error (RMSE) for the tests.
\begin{table}
\caption{Tracking RMSE by Leg [m]}
\label{tab:track_error}
\centering
\begin{tabular}{lcccc}
\toprule
 & \multicolumn{2}{c}{Payload Control} & \multicolumn{2}{c}{ArduPilot Baseline} \\
\cmidrule(lr){2-3}\cmidrule(lr){4-5}
Leg & Run 1 & Run 2 & Run 1 & Run 2 \\
\midrule
1 (S) & 0.72 & 0.62 & 0.82 & 0.86 \\
2 (W) & 0.70 & 0.70 & 0.74 & 0.76 \\
3 (N) & 0.66 & 0.57 & 0.67 & 0.56 \\
4 (E) & 0.66 & 0.67 & 0.68 & 0.69 \\
\midrule
Square & 0.68 & 0.64 & 0.73 & 0.72 \\
\midrule
\emph{Mean of 8 Legs $\pm$ Std.} \\
Total RMSE & \multicolumn{2}{c}{$0.66 \pm 0.05$} & \multicolumn{2}{c}{$0.72 \pm 0.10$} \\
Transit RMSE & \multicolumn{2}{c}{$0.70 \pm 0.07$} & \multicolumn{2}{c}{$0.88 \pm 0.17$} \\
Hover RMSE & \multicolumn{2}{c}{$0.63 \pm 0.05$} & \multicolumn{2}{c}{$0.55 \pm 0.07$} \\
Settled Peak-to-Peak RMSE & \multicolumn{2}{c}{$0.59 \pm 0.18$} & \multicolumn{2}{c}{$0.89 \pm 0.35$} \\
\bottomrule
\end{tabular}
\end{table}
Since the vehicle-referenced baseline is agnostic to the payload error, integral action accounts only for the drone error, therefore the payload experienced a damped oscillation about a point that lags behind the reference due to aerodynamic drag. As shown in Fig. \ref{fig:position}, the vehicle-referenced baseline then immediately closed the error gap at the leg-to-hover transition as momentum carried the payload over the waypoint.
This decrease was followed by an increase in error as the payload overshot the waypoint. Because the payload-aware controller had already removed the accumulated payload error, the payload sat at or near the reference at the leg-to-hover transition, resulting in immediate position error increase. The payload-aware controller would immediately try to fly the drone in the opposite direction of the payload movement at the waypoint. This actuation caused a large payload swing. Whereas, with the vehicle-referenced baseline, the drone tended to slide over the waypoint which helped to damp payload swing. For both controllers peak swing occurred during waypoint transitions where the payload-aware controller reached 12.97$^\circ$ and the baseline reached 12.74$^\circ$. The large peak swing of the payload-aware controller and natural damping in the ArduPilot baseline resulted in an overall lower payload tracking RMSE for the baseline during the 15 s hover. However, with the payload-aware controller, the payload settled at a much smaller position error. This result is shown as the settled peak-to-peak RMSE in Table \ref{tab:track_error} which takes the RMSE of the last 5.0~s of the 15~s hover corresponding to one full pendulum period for a 6.2~m tether.\par

When starting a new leg, both controllers are subjected to the same actuation limits and experienced a similar position error. However, the payload-aware controller typically started from a smaller position error, and the vehicle-referenced baseline carried the error not damped out during the 15~s hover time. The payload-aware controller reduced the mean per-leg tracking RMSE from 0.72~m to 0.66~m and standard deviation from 0.10~m to 0.05~m. With four flights the difference in means is comparable to the variation between legs, but the payload-aware controller was the more consistent of the two, with a worst leg of 0.72~m against 0.86~m for the baseline. When considering only the transit portion of the leg, the payload-aware controller reduced the position RMSE from 0.88~m to 0.70~m, about 20\%, and, as shown in Table \ref{tab:track_error}, put the transit RMSE just outside of the baseline's standard deviation.\par

\begin{figure}[t]
    \centering
    \vskip 5pt
    \includegraphics[width=\columnwidth]{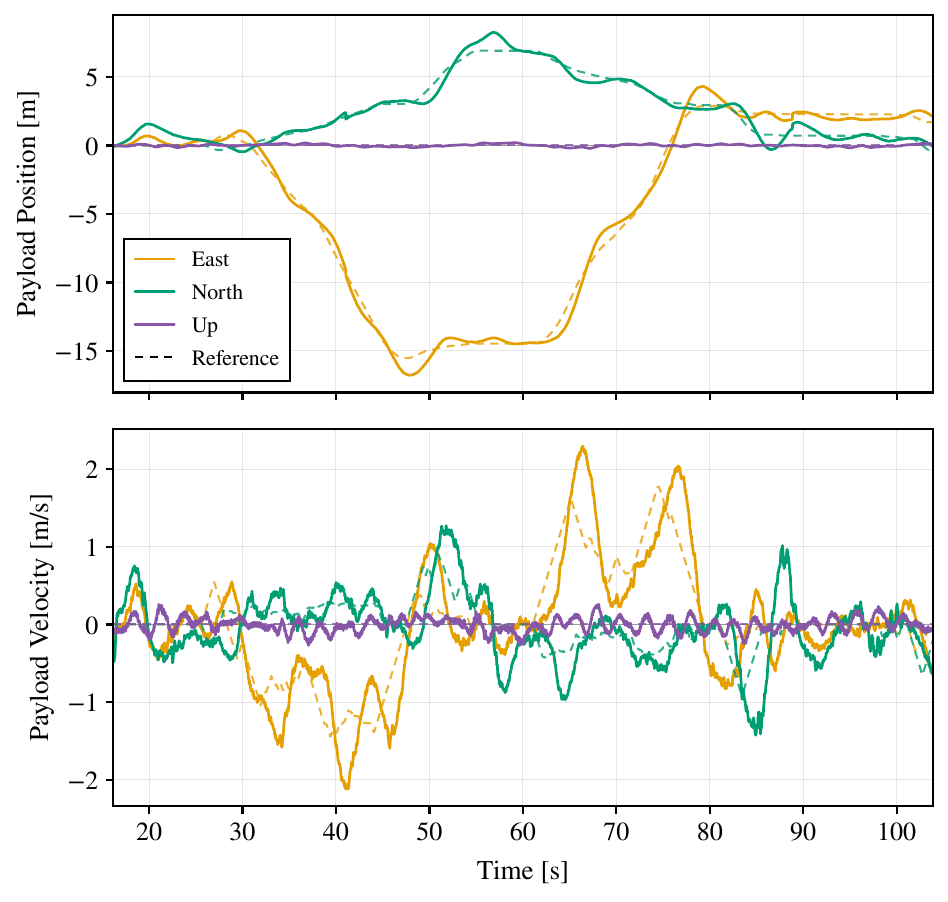}
    \caption{Payload position (top) and velocity (bottom) as commanded with the piloting aid. Orange, green, and purple show the East, North, and Up components respectively; dashed lines are the reference and solid lines the achieved response.}
    \label{fig:track}
\end{figure}

To demonstrate the piloting aid discussed in Sec. \ref{sec:pilot}, a human pilot flew a short loop with the maximum velocity set to 3 m/s and the maximum acceleration set to 0.4 m/s$^2$. In practice, the pilot prefers to have slow acceleration for fine tuning the payload position, but still requires fast speeds for longer range surveys. Fig. \ref{fig:track} contains the commanded velocity and the resulting reference position with time. The achieved position and velocity are overlaid.
When using the payload-aware control, the pilot would fly a leg, then come to a stop and change direction before the payload had settled. Waiting for the payload to come to a stop before moving again requires patience and battery life. This behavior motivated the slow velocity ramps, so that the pilot could return to motion without settling. In general, the pilot reported that payload-aware control was easier to fly. This mode is useful, since the pilot often wants to stop and dwell on a source, however, the reference trajectory generation could be further improved with smooth S-curves.\par

\section{Conclusion and Future Work}
In this work, we have designed an LQI acceleration controller for regulating the position of a tether-suspended payload. We have also implemented a vision-based method for estimating payload bearing by tracking a red ring, and a Kalman filter was designed to sustain estimates during brief dropouts. While the payload-aware controller struggled with larger swings at waypoint transitions, the controller reduced the payload tracking error in outdoor flight tests under wind when compared against the ArduPilot vehicle-referenced baseline.\par

The next step of this project is to take a radiation detecting payload to a real-world radioactive contamination survey. This survey is anticipated to have complex terrain contours and deep trenches that require the drone to precisely maneuver the payload. Maneuvers are expected to be slow and methodical, which should keep swing angles within the small-angle range assumed here, but failsafes remain necessary to ensure the system's safety. The control system developed now serves as the baseline for more advanced model-predictive control. The heavy-lift system is inherently more hazardous than lab-scale systems, so control advances must be added and validated carefully. With a more advanced control system, however, we will be able to perform faster and more agile surveys, leading to more intelligent autonomous systems deployed in the field.

\section{Acknowledgments}
This work was supported by the U.S. Army Corps of Engineers FUSRAP program. The authors thank the staff at Richmond Field Station for flight operations support, as well as August Lukkassen and Sahil Vazhathodiyil of the UC Berkeley HiPeRLab along with Wren Kawamura and Aidan O'Donnell of Lawrence Berkeley National Laboratory for field test support.

\bibliographystyle{IEEEtran}
\bibliography{references}

@article{aerial_sanada_2014, title={Aerial radiation monitoring around the Fukushima Dai-ichi nuclear power plant using an unmanned helicopter}, volume={139}, ISSN={0265-931X}, DOI={10.1016/j.jenvrad.2014.06.027}, abstractNote={The Great East Japan Earthquake on March 11, 2011 generated a series of large tsunami that seriously damaged the Fukushima Dai-ichi Nuclear Power Plant (FDNPP), which resulted in the release of radioactive materials into the environment. To provide further details regarding the distribution of air dose rate and the distribution of radioactive cesium (134Cs and 137Cs) deposition on the ground within a radius of approximately 5 km from the nuclear power plant, we carried out measurements using an unmanned helicopter equipped with a radiation detection system. The distribution of the air dose rate at a height of 1 m above the ground and the radioactive cesium deposition on the ground was calculated. Accordingly, the footprint of radioactive plumes that extended from the FDNPP was illustrated. - Highlights: • The radiation monitoring system using an unmanned helicopter was developed. • Monitoring and analysis methods using this system was established. • The reliability of this method was evaluated by good agreement with ground survey. • We carried out radiation survey of around the FDNPP using this system}, journal={Journal of Environmental Radioactivity}, author={Sanada, Yukihisa and Torii, Tatsuo}, year={2015}, month={Jan}, pages={p. 294–299} }

@book{probabilistic_thrun_2005,
	title = {Probabilistic Robotics},
	author = {Thrun, Sebastian and Burgard, Wolfram and Fox, Dieter},
	publisher = {MIT Press},
	address = {Cambridge, MA},
	series = {Intelligent Robotics and Autonomous Agents},
	year = {2005},
	isbn = {9780262201629}
}

@article{prototype_jiang_2016,
  title={A prototype of aerial radiation monitoring system using an unmanned helicopter mounting a GAGG scintillator Compton camera},
  author={Jianyong Jiang and Kenji Shimazoe and Yasuaki Nakamura and Hiroyuki Takahashi and Yoshiaki Shikaze and Yukiyasu Nishizawa and Mami Yoshida and Yukihisa Sanada and Tatsuo Torii and Masao Yoshino and Shigeki Ito and Takanori Endo and Kosuke Tsutsumi and Sho Kato and Hiroki Sato and Yoshiyuki Usuki and Shunsuke Kurosawa and Kei Kamada and Akira Yoshikawa},
  journal={Journal of Nuclear Science and Technology},
  year={2016},
  volume={53},
  pages={1067 - 1075},
  url={https://api.semanticscholar.org/CorpusID:101033993}
}

@article{remote_sato_2018,
author = {Yuki Sato and Shingo Ozawa and Yuta Terasaka and Masaaki Kaburagi and Yuta Tanifuji and Kuniaki Kawabata and Hiroko Nakamura Miyamura and Ryo Izumi and Toshikazu Suzuki and Tatsuo Torii},
title = {Remote radiation imaging system using a compact gamma-ray imager mounted on a multicopter drone},
journal = {Journal of Nuclear Science and Technology},
volume = {55},
number = {1},
pages = {90--96},
year = {2018},
publisher = {Taylor \& Francis},
doi = {10.1080/00223131.2017.1383211},


URL = { 
    
        https://doi.org/10.1080/00223131.2017.1383211
    
    

},
eprint = { 
    
        https://doi.org/10.1080/00223131.2017.1383211
    
    

}

}

@article{first_mochizuki_2017,
doi = {10.1088/1748-0221/12/11/P11014},
url = {https://doi.org/10.1088/1748-0221/12/11/P11014},
year = {2017},
month = {nov},
publisher = {},
volume = {12},
number = {11},
pages = {P11014},
author = {Mochizuki, S. and Kataoka, J. and Tagawa, L. and Iwamoto, Y. and Okochi, H. and Katsumi, N. and Kinno, S. and Arimoto, M. and Maruhashi, T. and Fujieda, K. and Kurihara, T. and Ohsuka, S.},
title = {First demonstration of aerial gamma-ray imaging using drone for prompt radiation survey in Fukushima},
journal = {Journal of Instrumentation}
}

@Article{radiation_towler_2012,
AUTHOR = {Towler, Jerry and Krawiec, Bryan and Kochersberger, Kevin},
TITLE = {Radiation Mapping in Post-Disaster Environments Using an Autonomous Helicopter},
JOURNAL = {Remote Sensing},
VOLUME = {4},
YEAR = {2012},
NUMBER = {7},
PAGES = {1995--2015},
URL = {https://www.mdpi.com/2072-4292/4/7/1995},
ISSN = {2072-4292},
DOI = {10.3390/rs4071995}
}

@inproceedings{dynamics_sreenath_2013,
  title={Dynamics, Control and Planning for Cooperative Manipulation of Payloads Suspended by Cables from Multiple Quadrotor Robots},
  author={Koushil Sreenath and Vijay R. Kumar},
  booktitle={Robotics: Science and Systems Conference},
  year={2013},
  url={https://api.semanticscholar.org/CorpusID:9105307}
}

@ARTICLE{circle_kasa_1976,
  author={Kåsa, I.},
  journal={IEEE Transactions on Instrumentation and Measurement}, 
  title={A circle fitting procedure and its error analysis}, 
  year={1976},
  volume={IM-25},
  number={1},
  pages={8-14},
  doi={10.1109/TIM.1976.6312298}}

@article{suspended_klausen_2014,
author = {Klausen, Kristian and Fossen, Thor and Johansen, Tor},
year = {2017},
month = {12},
pages = {},
title = {Nonlinear Control with Swing Damping of a Multirotor UAV with Suspended Load},
volume = {88},
journal = {Journal of Intelligent \& Robotic Systems},
doi = {10.1007/s10846-017-0509-6}
}

@article{foehn_fast_2017,
  title={Fast Trajectory Optimization for Agile Quadrotor Maneuvers with a Cable-Suspended Payload},
  author={Philipp Foehn and Davide Falanga and Naveen Suresh Kuppuswamy and Russ Tedrake and Davide Scaramuzza},
  journal={Robotics: Science and Systems XIII},
  year={2017}
}

@book{bryson1975applied,
  title     = {Applied Optimal Control: Optimization, Estimation, and Control},
  author    = {Bryson, Arthur E. and Ho, Yu-Chi},
  year      = {1975},
  edition   = {Revised},
  publisher = {Hemisphere Publishing},
  address   = {Washington, DC}
}

@Article{limiting_ravehamit_2022,
AUTHOR = {Raveh-Amit, Hadas and Sharon, Avi and Katra, Itzhak and Stilman, Terry and Serre, Shannon and Archer, John and Magnuson, Matthew},
TITLE = {Limiting Wind-Induced Resuspension of Radioactively Contaminated Particles to Enhance First Responder, Early Phase Worker and Public Safety—Part 1},
JOURNAL = {Applied Sciences},
VOLUME = {12},
YEAR = {2022},
NUMBER = {5},
ARTICLE-NUMBER = {2463},
URL = {https://www.mdpi.com/2076-3417/12/5/2463},
ISSN = {2076-3417},
DOI = {10.3390/app12052463}
}

@inproceedings{dynamic_gassner_2017,
author = {Gassner, Michael and Cieslewski, Titus and Scaramuzza, Davide},
year = {2017},
month = {05},
pages = {},
title = {Dynamic Collaboration without Communication: Vision-Based Cable-Suspended Load Transport with Two Quadrotors},
doi = {10.1109/ICRA.2017.7989609}
}

@article{vision_slabber_2021,
  title={Vision-Based Control of an Unknown Suspended Payload with a Multirotor},
  author={Johan F.M. Slabber and Hendrik Willem Jordaan},
  journal={2021 IEEE/RSJ International Conference on Intelligent Robots and Systems (IROS)},
  year={2021},
  pages={4875-4880},
  url={https://api.semanticscholar.org/CorpusID:245264226}
}

@article{mixed_tang_2015,
author = {Tang, Sarah and Kumar, Vijay},
year = {2015},
month = {06},
pages = {2216-2222},
title = {Mixed Integer Quadratic Program trajectory generation for a quadrotor with a cable-suspended payload},
volume = {2015},
journal = {Proceedings - IEEE International Conference on Robotics and Automation},
doi = {10.1109/ICRA.2015.7139492}
}

@INPROCEEDINGS{swing_free_palunko,
  author={Palunko, Ivana and Fierro, Rafael and Cruz, Patricio},
  booktitle={2012 IEEE International Conference on Robotics and Automation}, 
  title={Trajectory generation for swing-free maneuvers of a quadrotor with suspended payload: A dynamic programming approach}, 
  year={2012},
  volume={},
  number={},
  pages={2691-2697},
  doi={10.1109/ICRA.2012.6225213}}

@article{trajectory_sreenath_2013,
  title={Trajectory generation and control of a quadrotor with a cable-suspended load - A differentially-flat hybrid system},
  author={Koushil Sreenath and Nathan Michael and Vijay R. Kumar},
  journal={2013 IEEE International Conference on Robotics and Automation},
  year={2013},
  pages={4888-4895}
}

@article{geometric_goodarzi_2015,
author = {Goodarzi, Farhad and Lee, Daewon and Lee, Taeyoung},
year = {2015},
month = {09},
pages = {1486-1498},
title = {Geometric control of a quadrotor UAV transporting a payload connected via flexible cable},
volume = {13},
journal = {International Journal of Control, Automation and Systems},
doi = {10.1007/s12555-014-0304-0}
}

@article{aggressive_tang_2018,
author = {Tang, Sarah and Wüest, Valentin and Kumar, Vijay},
year = {2018},
month = {01},
pages = {1152-1159},
title = {Aggressive Flight With Suspended Payloads Using Vision-Based Control},
volume = {3},
journal = {IEEE Robotics and Automation Letters},
doi = {10.1109/LRA.2018.2793305}
}

@article{pathfollowing_qian_2020,
author = {Qian, Longhao and Liu, Hugh},
year = {2020},
month = {03},
pages = {2021-2029},
title = {Path-Following Control of A Quadrotor UAV With A Cable-Suspended Payload Under Wind Disturbances},
volume = {67},
journal = {IEEE Transactions on Industrial Electronics},
doi = {10.1109/TIE.2019.2905811}
}

@article{modeling_gomiero_2024,
  author={Gomiero, Sara and von Ellenrieder, Karl D.},
  journal={IEEE Transactions on Automation Science and Engineering}, 
  title={Modeling and Sliding Mode Control of a Heavy-Lift Quadrotor With a Cable-Suspended Payload Under Wind Disturbances}, 
  year={2026},
  volume={23},
  number={},
  pages={3065-3082},
  doi={10.1109/TASE.2025.3612189}}

@article{survey_villa_2020,
author = {Villa, Daniel and Brandão, Alexandre Santos and Sarcinelli-Filho, Mário},
year = {2020},
month = {05},
pages = {1-30},
title = {A Survey on Load Transportation Using Multirotor UAVs},
volume = {98},
journal = {Journal of Intelligent \& Robotic Systems},
doi = {10.1007/s10846-019-01088-w}
}

@Article{review_estvez_2024,
AUTHOR = {Estevez, Julian and Garate, Gorka and Lopez-Guede, Jose Manuel and Larrea, Mikel},
TITLE = {Review of Aerial Transportation of Suspended-Cable Payloads with Quadrotors},
JOURNAL = {Drones},
VOLUME = {8},
YEAR = {2024},
NUMBER = {2},
ARTICLE-NUMBER = {35},
URL = {https://www.mdpi.com/2504-446X/8/2/35},
ISSN = {2504-446X},
DOI = {10.3390/drones8020035}
}

\end{document}